# Survival of the Fittest in PlayerUnknown's BattleGrounds

Brij Rokad
Institute of Artificial Intelligence
brijrokad@uga.edu

Tushar Karumudi
Department of Computer Science
tushar.karumudi@uga.edu

Omkar Acharya
Department of Computer Science
oma52562@uga.edu

Akshay Jagtap
Department of Computer Science
akshay.jagtap@uga.edu

**Abstract**

The goal of this paper was to predict the placement in the multiplayer game PUBG (playerunknown's battlegrounds). In the game, up to one hundred players parachutes onto an island and scavenge for weapons and equipment to kill others, while avoiding getting killed themselves. The available safe area of the game's map decreases in size over time, directing surviving players into tighter areas to force encounters. The last player or team standing wins the round. In this paper specifically, we have tried to predict the placement of the player in the ultimate survival test. The data set has been taken from Kaggle. Entire dataset has 29 attributes which are categories to 1 label(winPlacePerc), training set has 4.5 million instances and testing set has 1.9 million. winPlacePerc is continuous category, which makes it harder to predict the "survival of the fittest." To overcome this problem, we have applied multiple machine learning models to find the optimum prediction. Model consists of LightGBM Regression (Light Gradient Boosting Machine Regression), MultiLayer Perceptron, M5P (improvement on C4.5) and Random Forest. To measure the error rate, Mean Absolute Error has been used. With the final prediction we have achieved MAE of 0.02047, 0.065, 0.0592 and 0634 respectively.

## 1. Introduction

*Battlegrounds* is a player versus player shooter game in which up to one hundred players fight in a battle royal, a type of large-scale last man standing deathmatch where players fight to remain the last alive. Players can choose to enter the match solo, duo, or with a small team of up to four people. There will be total of 100 players and the last person or team alive wins the match.[5]

Each match starts with players parachuting from a plane onto one of the four maps, with areas of approximately 8 × 8 kilometers (5.0 × 5.0 mi), 6 × 6 kilometers (3.7 × 3.7 mi), and 4 × 4 kilometers (2.5 × 2.5 mi) in size. The plane's flight path across the map varies with each round, requiring players to quickly determine the best time to eject and parachute to the ground. Players start with no gear beyond customized clothing selections which do not affect gameplay. Once they land, players can search buildings, ghost towns and other sites to find weapons, vehicles, armor, and other equipment. These items are procedurally distributed throughout the map at the start of a match, with certain high-risk zones typically having better equipment. Killed players can be looted to acquire their gear as well. Players can opt to play either from the first-person or third-person perspective, each having their own advantages and disadvantages in combat and situational awareness; though server-specific settings can be used to force all players into one perspective to eliminate some advantages.

Every few minutes, the playable area of the map begins to shrink down towards a random location, with any player caught outside the safe area taking damage incrementally, and eventually being eliminated if the safe zone is not entered in time; in game, the players see the boundary as a shimmering blue wall that contracts over time. This results in a more confined map, in turn increasing the chances of encounters. During the course of the match, random regions of the map are highlighted in red and bombed, posing a threat to players who remain in that area. In both cases, players are warned a few minutes before these events, giving them time to relocate to safety. A plane will fly over various parts of the playable map occasionally at random, or wherever a player uses a flare gun, and drop a loot package, containing items which are typically unobtainable during normal gameplay. These packages emit highly visible red smoke, drawing interested players near it and creating further confrontations. On average, a full round takes no more than 30 minutes. Considering that this game among the top 5 games sold worldwide and a recipient of the best multiplayer game of the year, it sparked an interest for us to estimate the placement/ survivability of a player in a game. We used multiple machine learning models to understand the same using the data of over a million players that was acquired from a public dataset. Since the output is a continuous variable we inclined to use of random forest, multilayer perceptron, light gradient boosting model, and M5P.

The next section of this paper will dive into the relevant works section citing the research that are related to this project. The following sections will explain our methods with the different machine learning algorithms that were used, and the results got from each of them.

## 2. Related Works:

The authors in [1] have discussed about the correlation-based feature selection. The authors argue that a good feature set contains of features that are highly correlated with the class, yet uncorrelated with each other. CFS (Correlation based Feature Selection) is an algorithm that couples this evaluation formula with an appropriate correlation measure and a heuristic search strategy. They have shown using artificial datasets that CFS quickly identifies and screens irrelevant, redundant, and noisy features, and identifies relevant features as long as their relevance does not strongly depend on other features. CFS measures correlation between nominal features, so numeric features are first discretized. CFS operates on the original (albeit discretized) feature space, meaning that any knowledge induced by a learning algorithm, using features selected by CFS, can be interpreted in terms of the original features, not in terms of a transformed space. On the other hand, the authors in [2] have rather sorted an interesting problem in hand, by using a different branch of machine learning. Empirical learning, is concerned with building or revising models in the light of large numbers of exemplary cases, taking into account typical problems such as missing data and noise. M5 builds tree-based models but, whereas regression trees have values at their leaves, the trees constructed by M5 can have multivariate linear models; these model trees are thus analogous to piecewise linear functions. The aim of the authors is to construct a model that relates the target values of the training cases to their values of the other attributes. Tree-based models are constructed by the divide-and-conquer method. Unless Tree contains very few cases, or their values vary only slightly, T is split on the outcomes of a test. Every potential test is evaluated by determining the subset of cases associated with each outcome.

The authors in [3] have compared two of the methods we used the LGBM and the MLP. Both these methods have been effective on the continuous attribute classes. They assessed two state of the art gradient boosting methods, namely eXtreme Gradient Boosting (XGBoost) and Light Gradient Boosting Machine (LGBM) to predict the probability that a driver will initiate an auto insurance claim in the framework of the Kaggle Porto Seguro Challenge. The main idea here is to use LGBM or XGBoost feature importance method to reduce the dimensionality of the training data not by selecting the important ones but rather by eliminating the unimportant ones. The underlying idea is that if the gradient boosting splitting algorithm does not select some feature to split, it means that this feature does not add value to the boosting trees model. The results from the paper efficiently showcase that the gradient boosted trees can outperform the MLP. They argue that this method of gradient boosting can be applied to all types of datasets and can produce a similar result.

Similar to [3], the authors of [4] have tried to compare the predictions of M5P and MLP discussed above. Here the authors have tried predicting the prices of diamonds basing on the attributes they have from a dataset. They evaluated the performance of the models developed using various standard statistical performance evaluation criteria. for the author's dataset, the M5P model produced better overall results than linear regression and neural network when evaluated on tenfold cross validation. The authors claim when correlation or feature reduction is done the performance of M5P increases rapidly as oddly opposed to the performance of the MLP.

The authors in [7] have explained about the generating decision trees using the entropy minimization heuristic for discretizing the continuous-valued attributes. The technique uses the threshold value which they call it as the "cut point", in order to partition into intervals in the ID3. The discretization algorithms are used in the any algorithm which is working towards classification tree. They adopt the top-down **decision tree** generation like that of the ID3 and then give the entropy minimization for selection of an attribute which is related to the feature selection process in our project. Here the authors compute the entropy for each data partition which gives the amount of information in bits which is given randomly and smaller the entropy value, the less even is the class distribution, to overcome this, they use the weighted average of the resultant class entropies. Therefore, the efficiency is greatly differed when it comes to cut point selection, so the use of the classification algorithm can be used to compute the information entropy minimization heuristic for selecting best cut points. While, authors in [9], describe the ensemble classification and regression approach for benchmarking result in prediction error computation and node splits with tuning the tree size and limiting the number of splits, using the same settings, they have mentioned that RandomForest gives good performance even with larger primary tuning values. The authors use the Least Angle Regression (LAR) to include a special case of penalizing the regression and also guards from overfitting. The author applies random forest regression on the mlbench datasets which generates coinciding result with bagging. The prediction error obtained by the suite of RandomForest gives the optimal single split, but the authors did not the test same settings of the RandomForest on larger numbers with very limited significant inputs. Now comparing with the [9], here the authors have used the unsupervised learning methods for feature extraction from unlabelled data. Unlike the supervised algorithms (RandomForest), authors have compared the results with the auto-encoders and K-Means on the ImageNet Images labelled dataset, where the algorithms used by authors takes in the random guesses and linear filters tied up with each neuron for classifying the face againsts the distractors. The author got very less accuracy for the random guess with the unlabelled ImageNet dataset with different data splits at every validation but with random splits at the training stage, the results vary largely as compared to previous results, this leads to increment of fine-tuning to the entire network for classification.

## 3. Data Analysis:

Kaggle has provided with a large number of anonymized PUBG game stats, formatted so that each row contains one player's post-game stats. The data comes from matches of all types: solos, duos and squads. Data set is divided into two parts: training set and testing set. Training set contains 4.5 million instances with 29 attributes, whereas testing set contains 1.9 million instances with 28 attributes (without label). With this much big dataset, there is bound to have some redundancies. For example, there might not be all 100 players in one game. Some games have less than 100 players. So, to begin with we have tried to do statistics on number of kills, we found out that most of the people have 0 kills. What about the damage being dealt by them, then? As expected, our analysis shows that most of the 0 kill players don't do damage at all, but sometimes in PUBG you can win the game even without killing anyone.

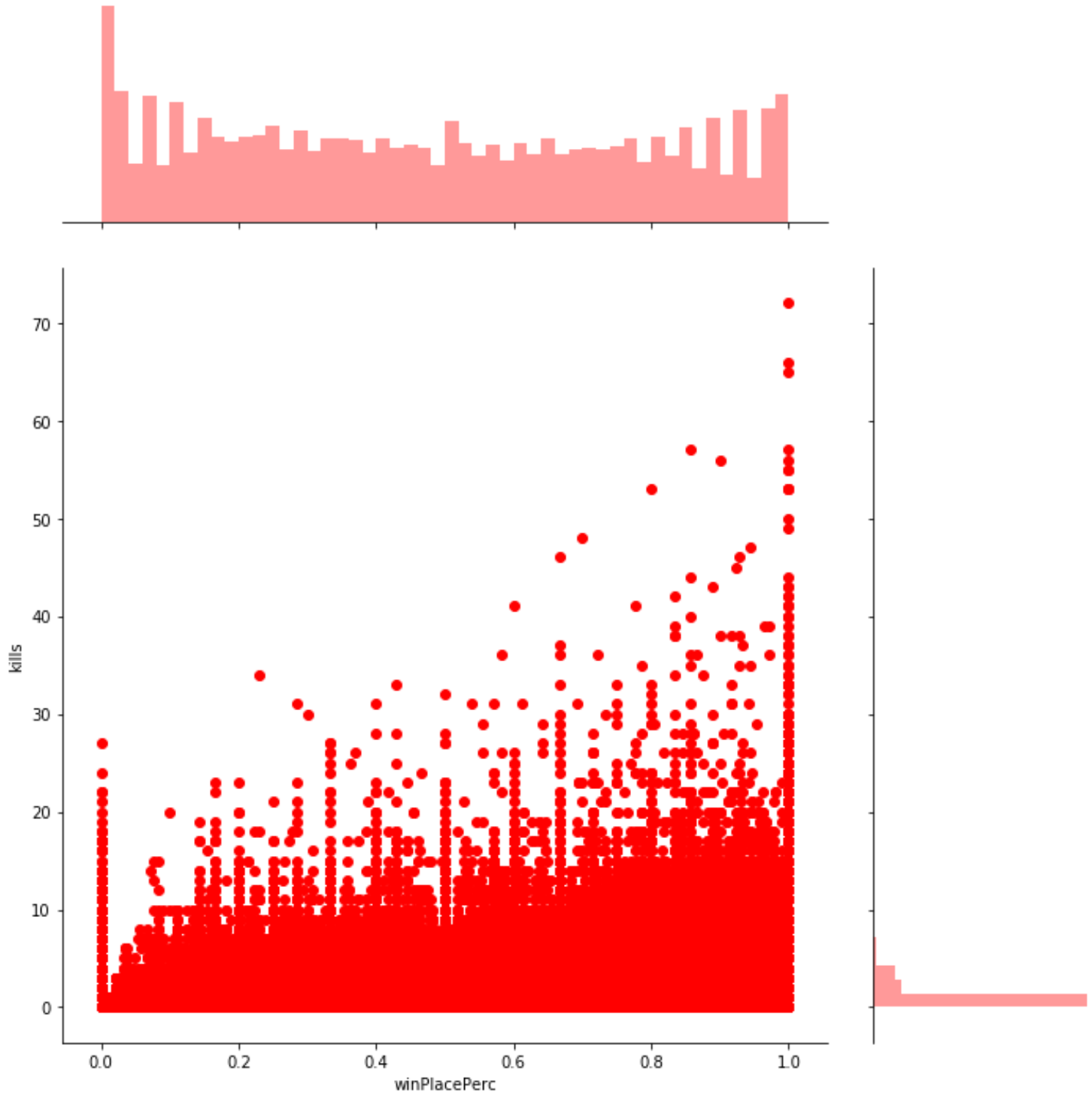

Figure 1: winPlacePerc VS kills

So, to get an insight of that, we have plotted some data points. Figure 1 shows the winPlacePerc VS kills and we can clearly see that only fraction of the total players are able to win without a single kill. Only 0.3748% of the total player have won without single kill out of which 0.1059% players have won without a kill and without dealing damage. It’s clear here that: higher your kills are, more chances for you to win.

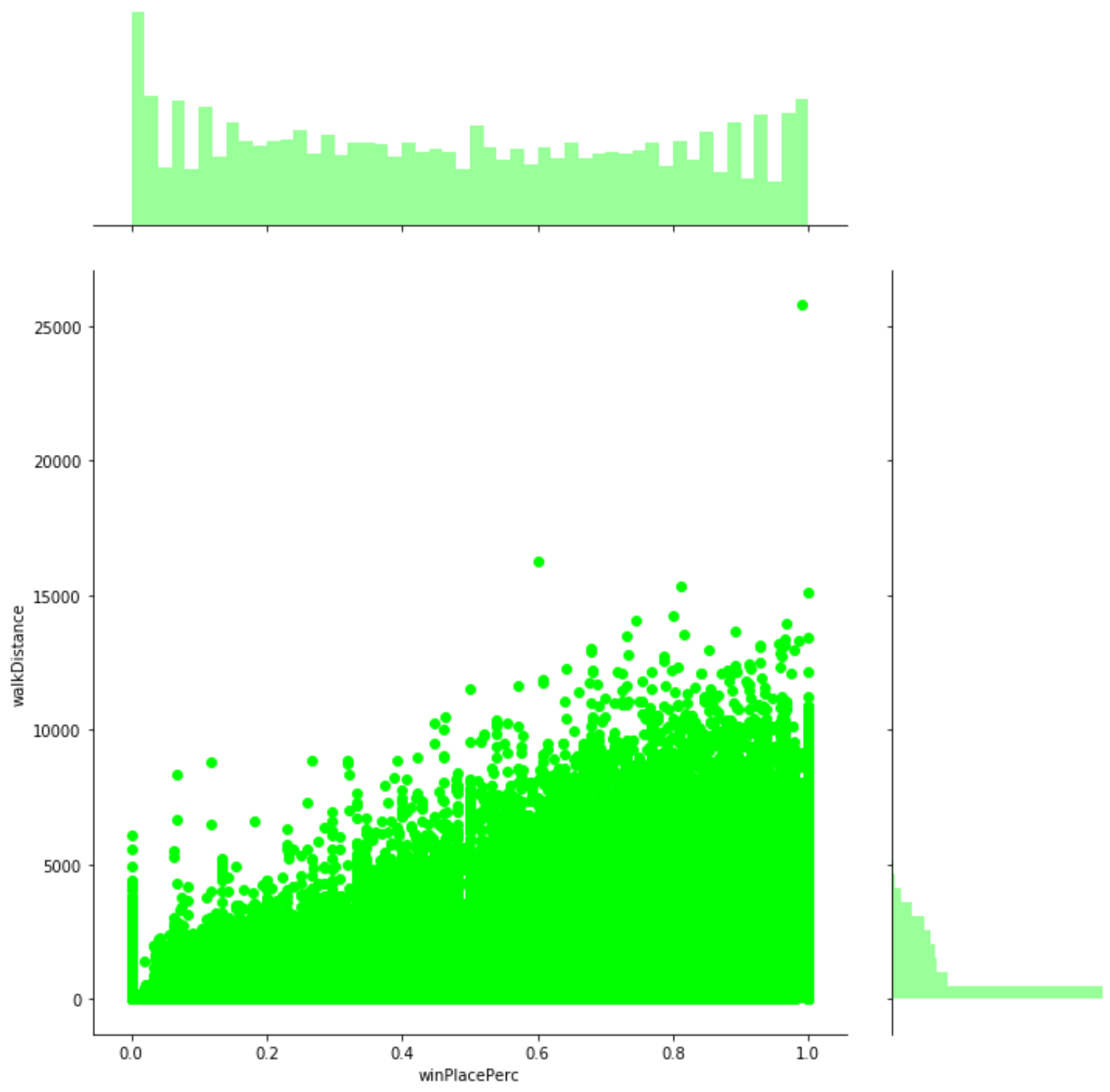

Figure 2: winPlacePerc VS walkDistace

But that’s not the end, because of very large map, most of the player travel with either by walk or by a vehicle. We have plotted a graph of winPlacePerc VS walkDistance, as shown in figure 2 plot clearly shows that the more you walk, more chances are there for you to win, apart from that our analysis showed us that 2.0329% players have died before even taking a single step. We call these types of players as AFK (away from keyboard) which will only act as redundant instances in our data set. We have removed all the redundant instances like AFK from the dataset to make it smoother.

In PUBG we know that the player can play solo, duo and in a squad. But how do these players play and perform in each match type? Does their behavior changes to different play style? And answer to all these questions is, yes. After doing some analysis on match type, we came to know that 15.95% players play solo, 74.10% players play in duo and 9.95% players play in squad. We have plotted these data point in figure 3, which is kills VS winPlacePerc for all match types (Solo, Duo and Squad). Plot shows that with fewer number of kills, players who prefers to play as a solo or as a duo have higher chances of winning the game compared to players who prefers to play in a squad. Moreover, from the figure 2 we can also deduce that players’ behavior changes when they play in a squad. Players seems to be more aggressive when they play in a squad match type, compare to solo and duo where players play more cautiously to avoid getting kill. That's because a player can be revived only in duo and squad by their teammates and because of that players are more aggressive when they play in a squad.

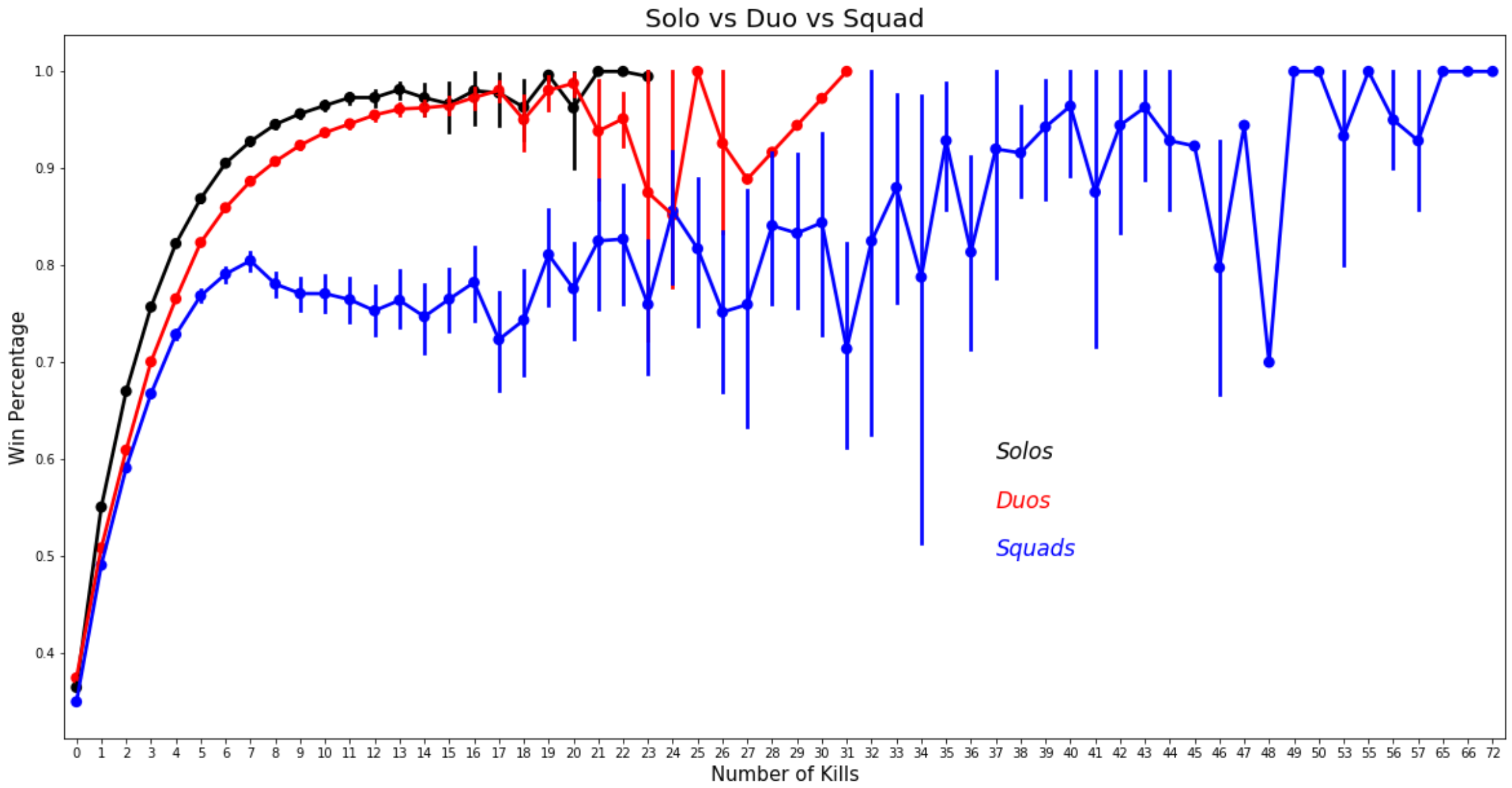

Figure 3: kills VS winPlacePerc for all type of matches. Figure also provides the necessary information to know the behaviour of players when they play in different types of match

Dataset has some unique attributes like Id, MatchId & GroupId which can be removed. Figure 3 shows correlation matrix among all attributes. From correlation matrix we got the Top 5/10 attributes which are highly correlated with label and low correlated with each other. Those 10 attributes are walkDistance, boosts, weaponAquired, damageDealt, heals, kills, longestkill, killStreaks, rideDistance & assist as shown in figure 4.

## 4. Feature Selection:

Our PUBG dataset had a total of 29 columns or predictors and our main focus was to filter out the columns which are not relevant to making prediction and keep only those predictors who are most significant. We have performed Feature Selection on our PUBG dataset using the "Select Attribute" in Weka Explorer. Feature Selection on Weka is mainly divided into two parts: Attribute Evaluator and Search Algorithm. Attribute Evaluator helps you to evaluate each feature from the dataset in context of output target class. Search Method assists you to checkout different combinations of the features from dataset in order to get the list of most significant predictors which have impact on the target class. We have used three Attribute evaluators namely CfsSubsetEval, ClassifierAttributeEval and Correlation Attribute Eval to get the best features for prediction.

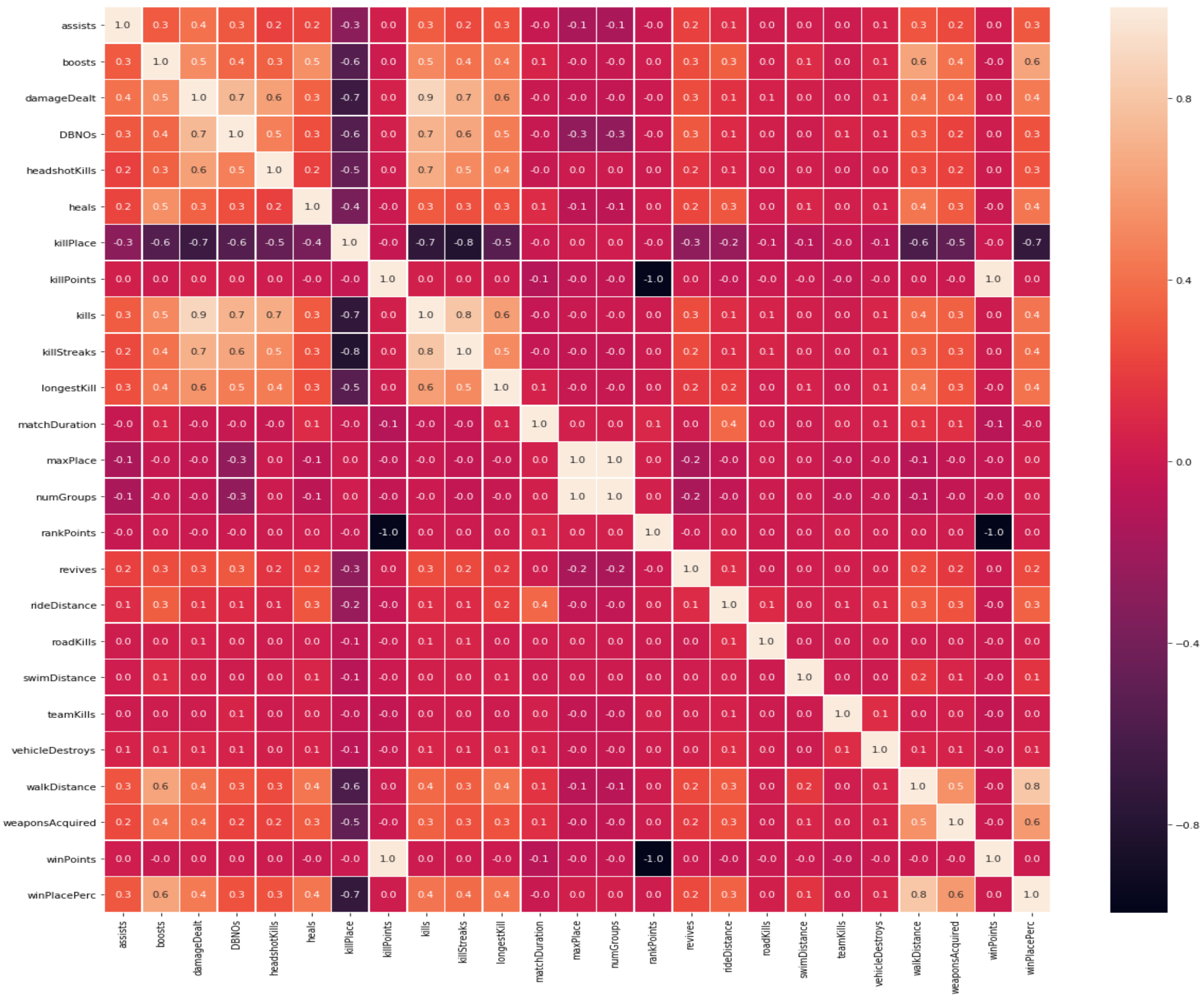

Figure 4: Correlation Matrix among all Attributes

CorrelationAttributeEval is a well-known technique used to do feature selection in Weka. Pearson's Correlation is basically a linear correlation between two variables X and Y. We calculate the correlation between attribute and target variable using the Ranker search method. We have selected only those attributes that have a moderately high positive or negative correlation and drop those columns which have values close to zero. Another popular technique to do feature selection is Information Gain. In this, we calculate the entropy for each attribute with respect to output variable and select only those features who have highest information gain. Fig.5 shows the result of applying CorrelationAttributeEval with Ranker search method on our dataset. As seen from Fig.5, CorrelationAttributeEval gave the list of attributes with the most significant attribute at the top with best rank. In this case, it was “walkDistance” feature with rank of 0.81119.

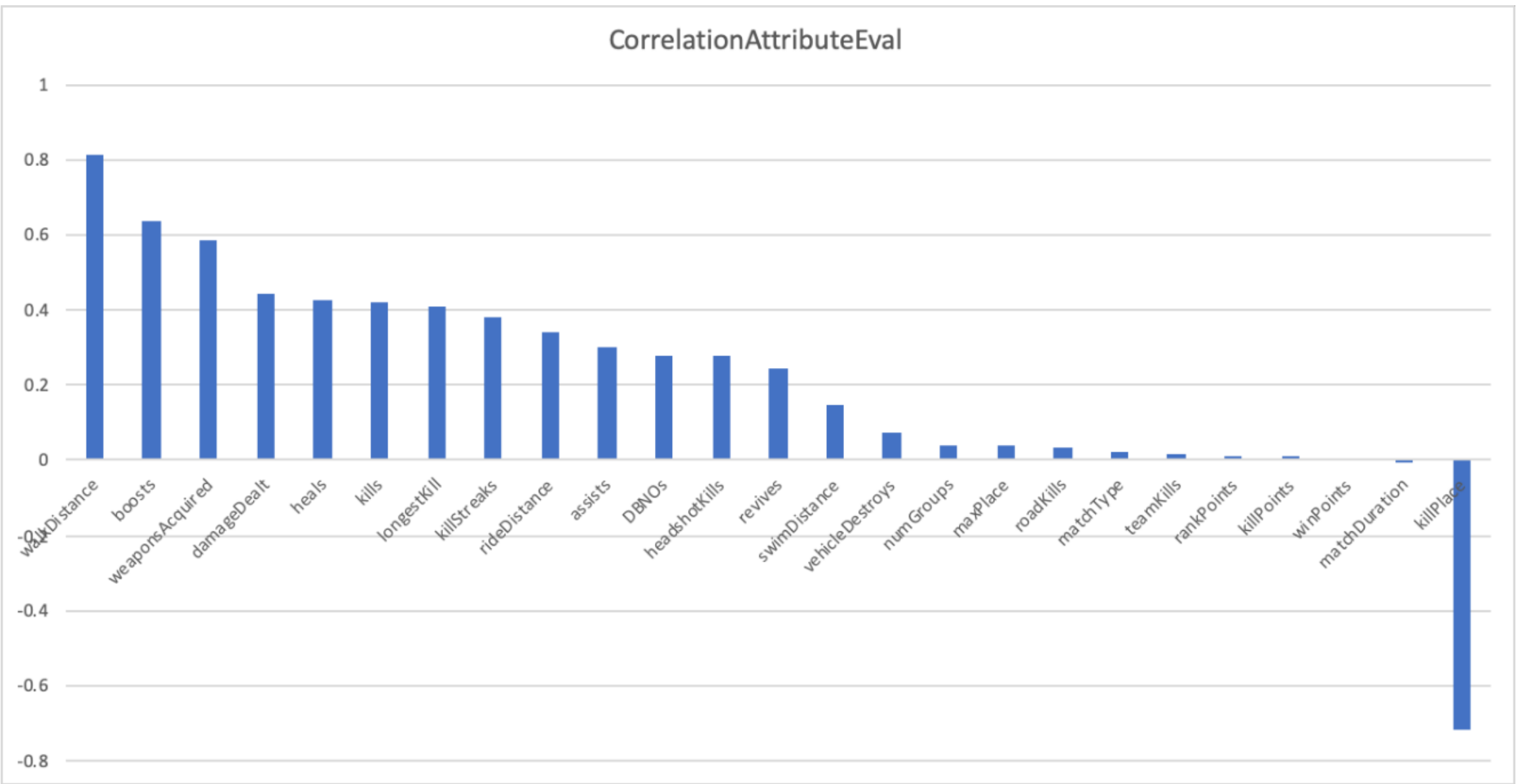

Figure 5: CorrelationAttributeEval based Feature Selection in Weka tool

For our PUBG dataset, the best attribute evaluator and the corresponding search method that gave most significant predictors was ClassifierAttributeEval using Ranker search method. The technique selected 18 attributes from total of 29 columns. The parameters used with this technique were: Folds: 5, threshold: 0.01 and classifier: M5P. Fig.6 shows the output after applying ClassifierAttributeEval.

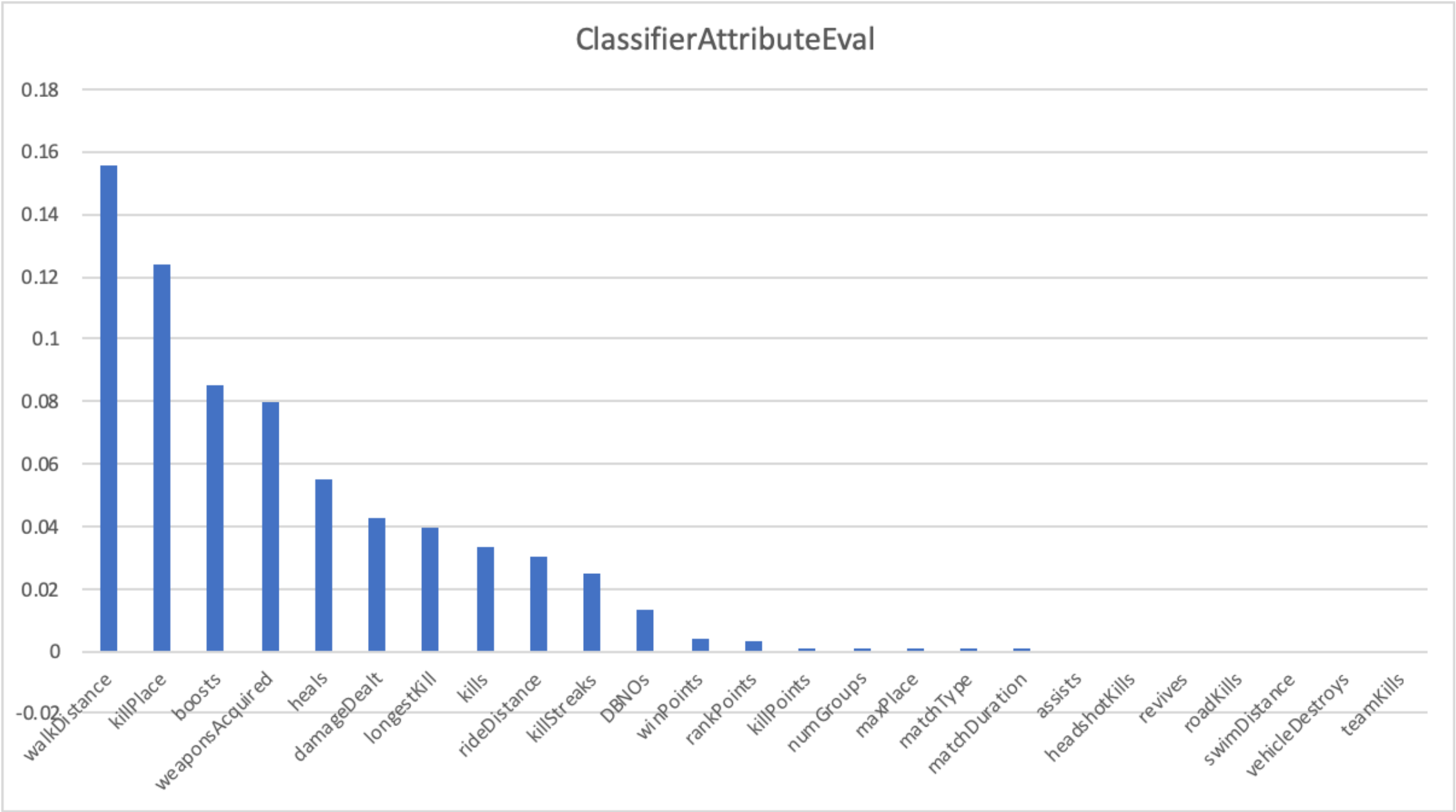

Figure 6: ClassifierAttributeEval based Feature Selection in Weka tool

As seen from Fig.6, the ClassifierAttributeEval gave a list with "walkDistance" as the attribute with most impact on target variable with a value of 0.150344. From the list, we removed all the attributes which had value close to zero. In this case, we removed assists, headshotkills, revives, roadkills, swimDistance, vehicleDestroys, teamkills and used rest of the columns for prediction. Also, we removed all the unique columns like Id, groupId and matchId from the dataset. Our prediction gave better results after removing above mentioned columns.

## 5. Feature Engineering

So, we have done feature selection on PUBG data set, but that wasn't enough for our data set. We tried to build some new features based on the current attributes. We have created 10 new features by normalizing the some of the current features. For example, we have normalized kills as killNorm and damageDealt as damageDealtNorm by taking into account of how many players have joined a particular match. When there are 100 players in the game it might be easier to find and kill someone, than when there are 90 players. So, we have normalized the kills in a way that a kill in 100 players will score 1 (as it is) and in 90 players it will score (100-90)/100 + 1 = 1.1. Apart from that healsAndBoosts and totalDistance also have been created for easier prediction.

New features created are: playerJoined, killsNorm, damageDealtNorm, healsAndBoosts, totalDistance, boostsPerWalkDistance, healsPerWalkDistance, killsPerWalkDistance, healsAndBoostsPerWalkDistance & team.

## 6. Methodology

### 6.1 Multilayer Perceptron

A multilayer perceptron (MLP) is a class of feedforward artificial neural network. The MLP consists of at least three layers of nodes: an input layer, a hidden layer and an output layer. Except for the input nodes, each node is a neuron that uses a nonlinear activation function. MLP utilizes a supervised learning technique called backpropagation for training. Its multiple layers and non-linear activation distinguish MLP from a linear perceptron. It can distinguish data that is not linearly separable. We trained the MLP in Weka. Multiple models were trained and tested to get the optimum combination of hidden nodes. Each combination was run for the same dataset size on 10-fold cross validation. The mean absolute error was noted for each of the combination. This step was repeated until we achieved the optimum combination. The total number of hidden layers present was 18 in the MLP. We set the learning rate to be 0.1 and the momentum also to be 0.1. The model trained using 10000 epochs. We used 10-fold cross validation to test the model built. 10-fold cross validation assures that each of the sample is tested at least once.

### 6.2 M5P

M5P is a Binary regression tree model where the last nodes are the linear regression functions that can produce continuous numerical attributes. Although we know that algorithms that generate decision trees are robust, efficient and relatively simple in classifying data, they do not perform well when we have a use-case of predicting continuous values. We have several other algorithms for predicting real values. Regression trees are one of the important algorithms for predicting real values that differ from decision trees only in having values than classes at the leaf level. So, for predicting real values, M5 is a new learning model that uses decision trees along with multivariate Linear regression. M5P builds decision trees and where we have regression trees predicting real values at leaves, M5 uses linear regression. Model trees are similar to piecewise linear function. This algorithm can efficiently handle many attributes/predictors from the dataset and the trees produced by M5 are much smaller when compared to its counterpart and are more accurate than regression trees. We trained our M5 model with the PUBG dataset on Weka tool. The number of instances used for training the model was 1 million records. To select the most significant predictor columns, we have used “ClassifierAttributeEval” attribute evaluator along with Ranker search algorithm given in Weka tool. We got 19 most important columns of our dataset which were used to train the model. The batchSize used for training was 100 and tree pruning was used to get optimized results. We tried different combinations of the model parameters but the best result we got was for batch Size of 100 with pruning and smoothing function enabled which helps to get better results for multivariate linear function running at the leaf level of the model tree. Also, 10-fold cross validation was used to ensure that each sample was tested, and we do not overfit the data.

### 6.3 Light Gradient Boosting Machine (LGBM Regression):

Light GBM is a gradient boosting framework that uses tree-based learning algorithm. So, what’s different about LGBM compare to other tree-based algorithms like Random Forest and Decision Making? Well in LGBM, algorithm grows tree vertically instead of horizontally whereas another tree-based algorithm grows tree vertically. In short, Light GBM grows tree **leaf-wise** while another algorithm grows level-wise. As shown in figure 7.

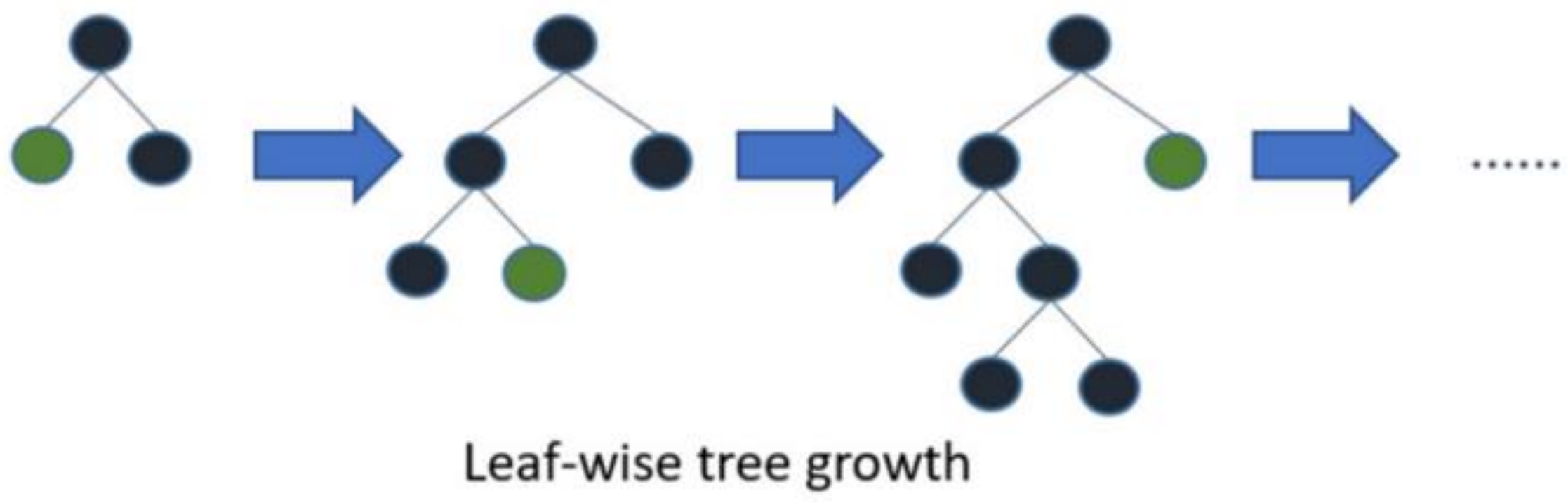

Figure 7: Light Gradient Boosting Machine

Light GBM will choose the leaf with max delta loss to grow. When growing the same leaf, Leaf-wise algorithm can reduce more loss than a level-wise algorithm. As the name suggest Light GBM is known for its high-speed predictions. Light GBM is useful to handle the large size of data and it also takes lower memory to run. The only drawback is that, it is sensitive to overfitting and can easily overfit small data.

Light GBM has more than 100 parameters which makes it complicated to implement and to tune all parameters. We have only tuned 8 parameters and rest of the parameters are set as default.

| | | |
|---|---|---|
| Core Parameters | objective: mae | To set the objective of the prediction |
| | num_leaves: 31 | Number of leaves in full tree |
| | random_state: 42 | To run models for different random samples. |
| Learning Control Parameters | learning_rate: 0.1 | Determines the impact of each tree on the outcome |
| | bagging_fraction: 0.7 | Specifies the fraction of data to be used for each iteration and is generally used to speed up the training and avoid overfitting |
| | feature_fraction: 0.7 | Used when boosting is gradient boosting, 0.7 feature fraction means LightGBM will select 70% of parameters randomly in each iteration for building trees |
| IO Parameters | verbose: 1 | IO parameter when the model fits |
| Matric | metric: mae | Metric to be evaluated on the evaluation set |

We have implemented Light GBM in python using scikit-learn in Kaggle Notebook, Kaggle notebook comes with high computational machines, which makes it easier to use entire dataset (4.5 training instances). We have built two different model, one with entire dataset (4.5 training instances) and another one with 1 million training instances so that we can compare the second model with other algorithms implemented in Weka.

**6.4 Random Forest:**

Random Forest algorithm is a supervised classification algorithm, creates the forest with several trees. The more trees in the forest the more robust the forest looks like. The higher the number of trees in the forest gives the high accuracy results. It can be used for the classification and regression task. It handles the missing values; the decision trees are more inclined towards the rule-based system. The set of rules will be used to perform prediction on the test dataset. The algorithm starts with selecting “k” features out of total “m” features and with the randomly selected “k” features, we find the root node by using the best split approach. Later, we calculate the child nodes using the same best split approach to reach the target leaf node. We repeat the same process and the resultant is the combination of multiple decision tree, thereby forming a random forest. As, it is form of decision tree, each internal node represents the outcome of the test and each branch represents a “test” on an attribute and every leaf node represents a class label. In Random Forest, we look for feature selections, which we can decide over the features to be dropped or add in the selected dataset. The advantage of Random Forest is that it won’t overfit the model, as there are enough trees which gives better prediction results. As compared to other decision tree classification algorithms, this algorithm gives high computational and quick to fit, even to the large problems along with handling of missing values using proximities. Even if we change the data a little, the individual trees may change but the forest is relatively stable because it is a combination of many trees, thereby giving the added advantage of instability, therefore justifying the algorithm robustness. Using the Random Forest algorithm, we trained the PUBG dataset in Weka tool. Initially, we tried with 1 million records, also performed “CorrelationAttributeEval” feature selection method which gave us the significant columns for prediction. Using the 19 columns after applying the feature selection, we applied Random Forest model with the BagSizePercent value 100 and batch size used for training set was 100 and numIterations with 100 iterations with 0 as the maximum depth of the tree which is unlimited depth, the more depth, higher is the prediction accuracy.

## 7. Results & Discussion:

### 7.1 Evaluation Metric:

We have used Mean Absolute Error as our base performance metric. Since the predicting class is a continuous variable. MAE measures the average magnitude of the errors in a set of predictions, without considering their direction. It’s the average over the test sample of the absolute differences between prediction and actual observation where all individual differences have equal weight. The Mean absolute error is calculated by the formula 1.

$$\mathrm{MAE} = \frac{\sum_{i=1}^{n} |y_i - x_i|}{n} \quad \text{--- (1)}$$

The other performance metric we have used is that of root mean square error. RMSE is a quadratic scoring rule that also measures the average magnitude of the error. It’s the square root of the average of squared differences between prediction and actual observation. The RMSE is represented using the following formula 2.

$$\text{RMSE} = \sqrt{\frac{1}{n}\sum_{j=1}^{n}(y_j - \hat{y}_j)^2} \quad \text{--- (2)}$$

**7.2 Results:**

The performance of M5P, LGBM, MLP, Random Forest algorithms evaluated on tenfold cross validation is given below in graph below:

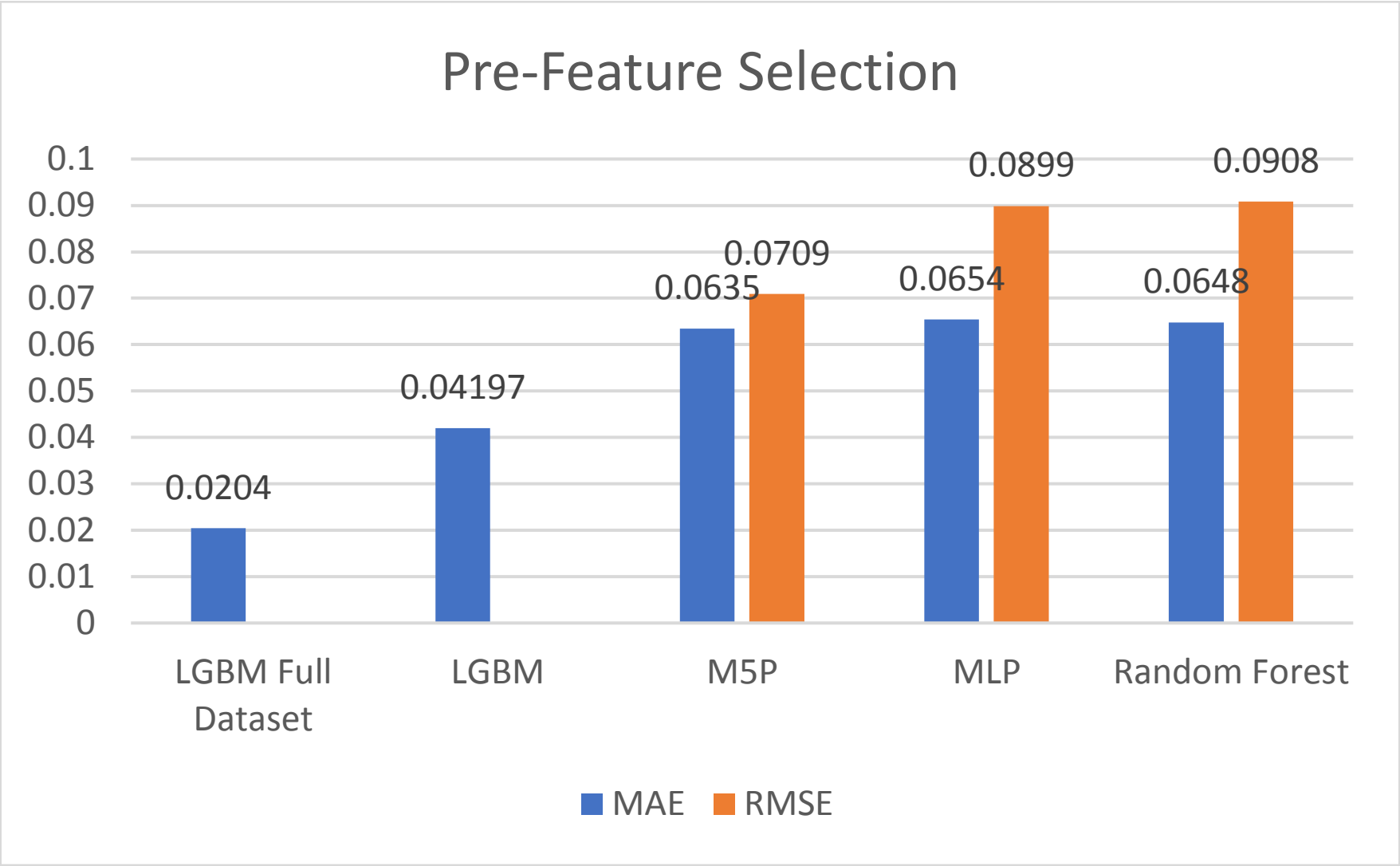

Figure 8: Pre-Feature Selection

The performance of the M5P, LGBM, MLP, Random Forest was evaluated after we performed the feature selection on the dataset. In the feature selection phase as we found some of the attributes were negatively affecting the performance or have low significance as identified in the stage above, we removed them and revaluated them. The results of performance of the algorithms post feature selection are:

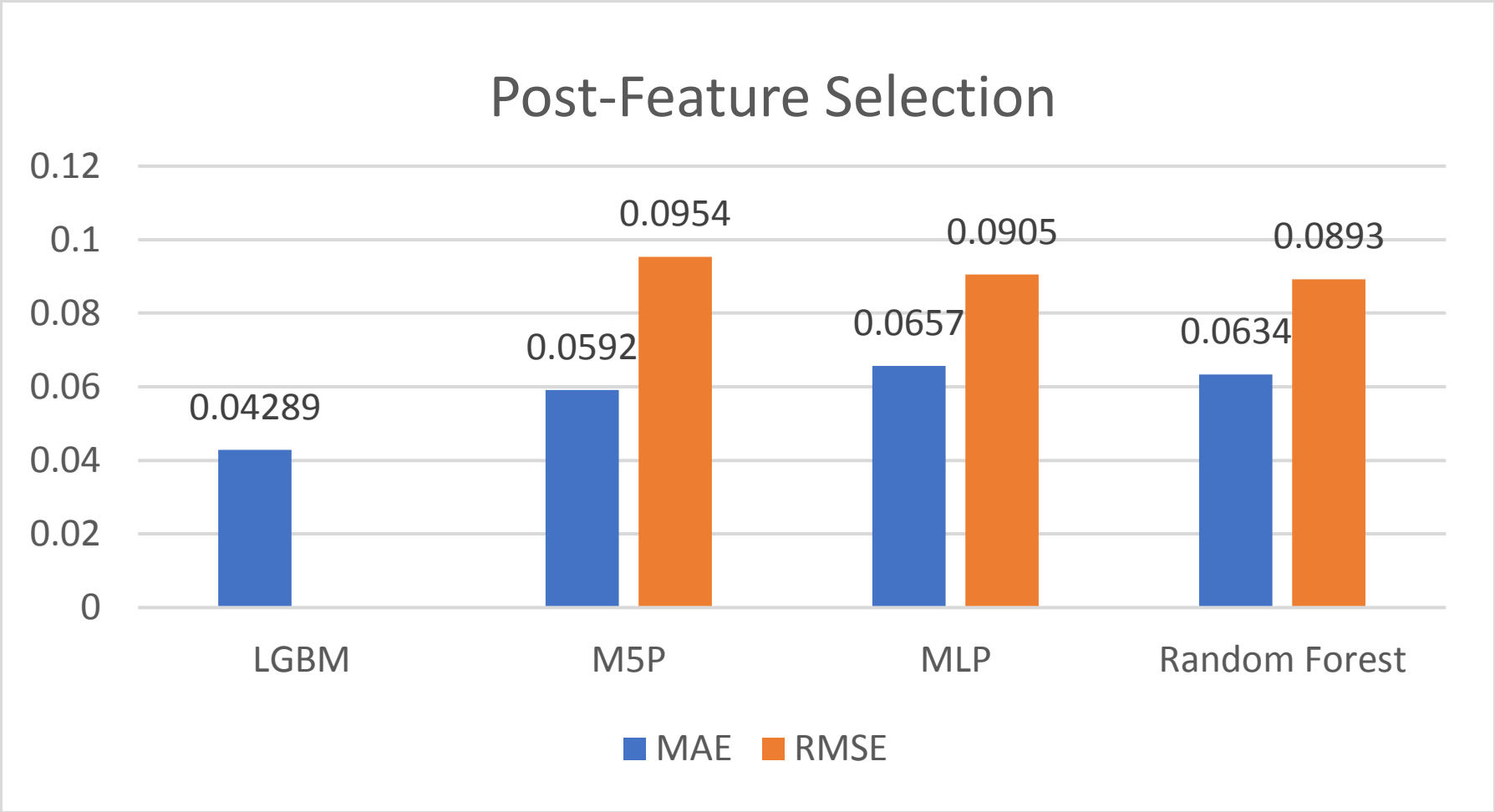

Figure 9: Post-Feature Selection

## 8. Conclusion:

From this study it can be concluded that machine learning techniques such MLP, Random Forest, LGBM and M5P model tree can be employed to predict the “survival of the fittest” in games like PUBG. Feature reduction by high correlation has proved to be a useful technique for performance improvement, although it might not apply for every scenario. Future work shall include more regression models in order to extend the robustness and precision of the predictions.